\documentclass[runningheads]{llncs}

\usepackage{eccv}

\usepackage{eccvabbrv}

\usepackage{graphicx}
\usepackage{booktabs}
\usepackage{multirow}
\usepackage{microtype}

\usepackage[accsupp]{axessibility}  

\usepackage{hyperref}

\usepackage{orcidlink}

\begin{document}
\raggedbottom

\title{G-ZAP: A Generalizable Zero-Shot Framework for Arbitrary-Scale Pansharpening}

\titlerunning{G-ZAP for Arbitrary-Scale Pansharpening}

\author{
Zhiqi Yang$^{*}$\orcidlink{0009-0001-3222-5883} \and
Shan Yin$^{*}$\orcidlink{0009-0004-4462-0559} \and
Jingze Liang\orcidlink{0009-0004-1925-9466} \and
Liang-Jian Deng$^{\dagger}$\orcidlink{0000-0003-3178-9772}
}

\authorrunning{Z.~Yang et al.}

\institute{
University of Electronic Science and Technology of China
}

\maketitle

\begingroup
\renewcommand{\thefootnote}{\fnsymbol{footnote}}
\footnotetext{$^{*}$The authors contributed equally. $^{\dagger}$Corresponding author.}
\endgroup

\begin{abstract}
Pansharpening aims to fuse a high-resolution panchromatic (PAN) image and a low-resolution multispectral (LRMS) image to produce a high-resolution multispectral (HRMS) image.
Recent deep models have achieved strong performance, yet they typically rely on large-scale pretraining and often generalize poorly to unseen real-world image pairs.
Prior zero-shot approaches improve real-scene generalization but require per-image optimization, hindering weight reuse, and the above methods are usually limited to a fixed scale.
To address this issue, we propose G-ZAP, a generalizable zero-shot framework for arbitrary-scale pansharpening, designed to handle cross-resolution, cross-scene, and cross-sensor generalization.
G-ZAP adopts a feature-based implicit neural representation (INR) fusion network as the backbone and introduces a multi-scale, semi-supervised training scheme to enable robust generalization.
Extensive experiments on multiple real-world datasets show that G-ZAP achieves state-of-the-art results under PAN-scale fusion in both visual quality and quantitative metrics.
Notably, G-ZAP supports weight reuse across image pairs while maintaining competitiveness with per-pair retraining, demonstrating strong potential for efficient real-world deployment.
\keywords{Pansharpening \and Zero-shot learning \and Arbitrary scale \and Implicit neural representation \and Generalization}
\end{abstract}

\section{Introduction}

\label{sec:intro}

\begin{figure}[ht]
    \centering
    \includegraphics[width=0.95\linewidth]{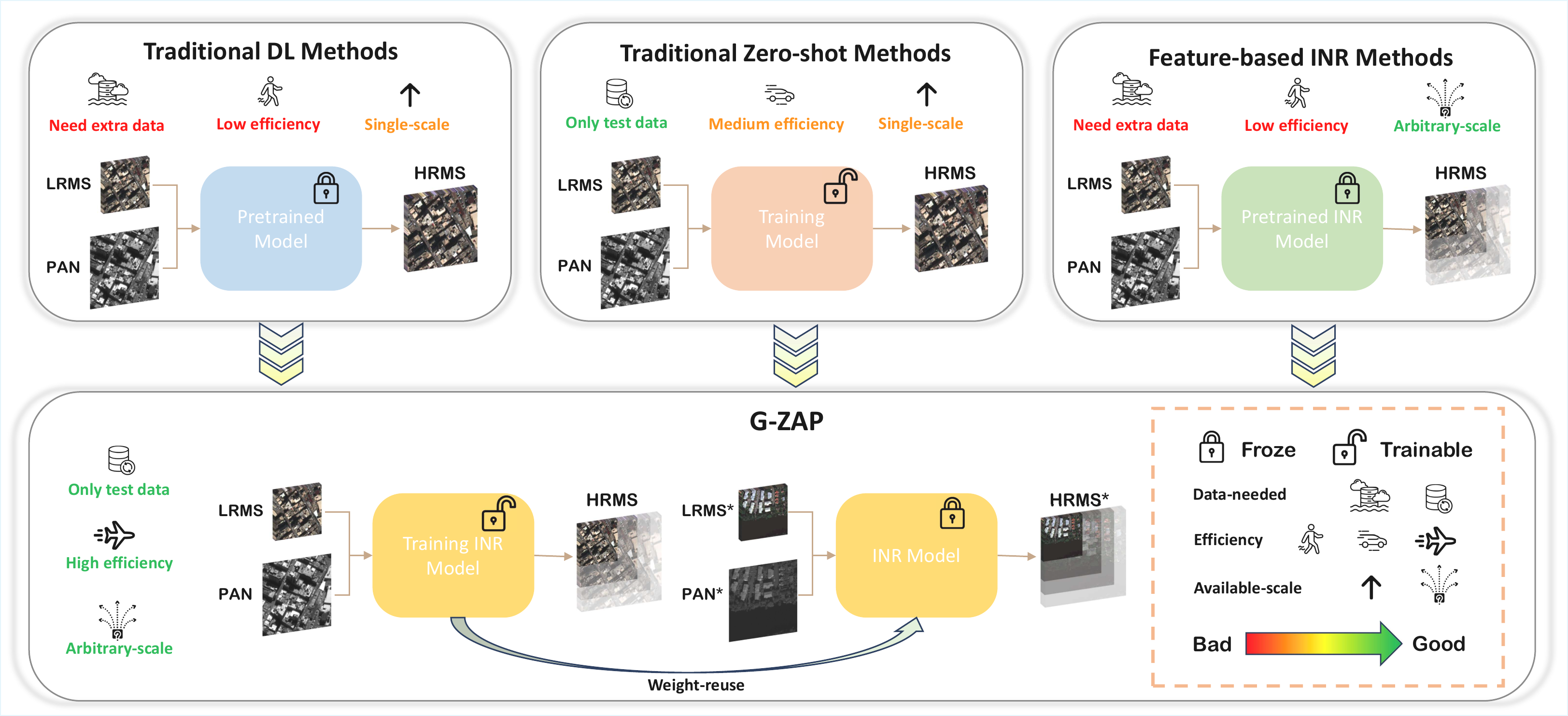}
    \caption{Comparison of traditional deep learning methods, traditional zero-shot methods, feature-based INR methods, and the proposed G-ZAP framework in terms of data requirement, efficiency, and available scale. G-ZAP achieves arbitrary-scale pansharpening with high efficiency using only test data through an INR-based zero-shot training strategy.}
    \label{fig:head}
\end{figure}

High-resolution multispectral remote sensing imagery plays a crucial role in fields such as urban planning, precision agriculture, environmental monitoring, and defense intelligence \cite{vivone2020new,deng2022machine,vivone2024deep,li2020review}. However, due to hardware limitations, it is often impractical to directly acquire multispectral images with both high spatial and high spectral resolution. Consequently, modern satellites typically capture two complementary types of images: low-resolution multispectral (LRMS) images and high-resolution panchromatic (PAN) images.
Pansharpening aims to fuse the LRMS and PAN images to generate a high-resolution multispectral (HRMS) image, thereby enhancing both the spatial and spectral details of the fused imagery.

Over the past several decades, pansharpening techniques have experienced significant advancements, evolving from classical model-driven approaches to data-driven deep learning paradigms. Early pansharpening methods mainly include component substitution (CS) \cite{kwarteng1989extracting,choi2010new}, multi-resolution analysis (MRA) \cite{vivone2013contrast,mtf-glp-fs}, and variational optimization (VO) \cite{fu2019variational,tian2021variational}. These traditional approaches are based on well-established mathematical models and aim to integrate spatial and spectral information without relying on training data. However, they often suffer from limitations such as spectral distortion and spatial inconsistency.
With the continuous progress in computational hardware and learning frameworks, deep learning-based pansharpening methods have shown remarkable potential in addressing these challenges, achieving superior performance in preserving both spatial details and spectral fidelity \cite{pnn,dicnn,pannet,lagconv,fusionmamba,wfanet,cao2025efficient,wu2025fully}.

However, most existing deep learning-based pansharpening methods are trained on simulated reduced-resolution datasets, which not only require large amounts of external training data but also lead to feature distortion when deployed at full resolution. Consequently, their performance on real-world datasets often falls short of that achieved on simulated benchmarks \cite{9745494}. To address the training-testing distribution mismatch, zero-shot pansharpening methods \cite{psdip,zspan} have recently gained increasing attention, as they learn models from scratch using only a single pair of input images and thus alleviate generalization issues. Nevertheless, these methods still suffer from several critical limitations. Specifically, they typically require image-specific optimization for each input pair, making retraining unavoidable when encountering new data and resulting in low efficiency for practical applications. Moreover, many prior approaches depend on accurate estimation of spatial fidelity to properly balance spectral fidelity, which increases methodological complexity and computational overhead. In addition, Pansharpening at arbitrary resolutions has great value in practical applications \cite{arhs}, but most existing methods are restricted to generating HRMS images at a fixed resolution corresponding to the size of PAN image, offering limited flexibility to meet diverse resolution requirements and potentially leading to insufficient clarity in certain details. 

Feature-based implicit neural representations (INRs) have recently attracted significant attention due to their remarkable capability in modeling continuous signals, where a shared function space is used to implicitly represent different objects. Such representations have been successfully applied to a variety of signal reconstruction tasks. For example, LIIF \cite{liif} and its variants \cite{liifv1,INR2,liifv3} learn continuous image representations to enable image super-resolution at arbitrary spatial resolutions. Similarly, methods such as INF3 \cite{infff} and FeINFN \cite{feinfn} leverage INR-based formulations to represent two-dimensional image signals for high-quality image fusion. However, existing feature-based INR image fusion approaches typically adopt a pretraining-testing paradigm to achieve arbitrary spatial resolution. To simultaneously address the aforementioned limitations and enable arbitrary-scale pansharpening, our work employs an feature-based INR image fusion model as the core framework and further investigates its performance in real-world pansharpening under a zero-shot setting.

To address the above issues, we propose a novel INR-centric Generalizable Zero-Shot Framework for Arbitrary-Scale Pansharpening (G-ZAP). The proposed method is built upon a three-level collaborative training framework.
Level-0 is designed to preserve features at full spatial resolution. Level-1 learns the standard 4\texttimes upsampling capability at one degraded resolution level, where a constructed ground truth is used as supervision to enhance the model's learning capacity. Building upon Level-1, Level-2 further introduces resolution degradation with varying scale factors, enabling the model not only to supervise the 4\texttimes scaling learned at Level-1, but also to learn a 16\texttimes scaling relationship with respect to Level-0.
By integrating unsupervised losses with constructed supervised losses, together with the implicit neural representation (INR)-based fusion model, the proposed G-ZAP framework is expected to simultaneously achieve strong cross-resolution, cross-scene, and cross-sensor generalization.
\cref{fig:head} compares traditional deep learning-based methods, traditional zero-shot strategies, feature-based INR approaches, and the proposed G-ZAP framework in terms of data requirement, efficiency, and available scale.

Overall, the contributions of this work can be summarized as follows:
\begin{enumerate}
  \item To the best of our knowledge, we are the first to integrate a feature-enhanced implicit neural representation INR-based fusion network into zero-shot pansharpening, enabling instant arbitrary-scale super-resolution for any single image pair without the need for pre-training.
  \item We propose a novel three-level training framework, termed G-ZAP, which achieves strong cross-resolution generalization. Extensive experiments demonstrate that the INR model trained on a single image pair can be directly transferred to unseen image pairs with negligible performance degradation, exhibiting robust cross-scene and cross-sensor generalization.
  \item Extensive experiments on multiple real-world remote sensing datasets show that the proposed method achieves state-of-the-art performance, and further demonstrate that superior results can be obtained without relying on spatial fidelity constraints under the proposed framework.
\end{enumerate}
\section{Related Work}

\subsection{Zero-Shot Learning}

In this paper, Zero-shot learning refers to a paradigm in which the model is trained without any pre-training, and both training and inference are conducted entirely at test time, aiming to learn from a single test sample or even zero samples. Shocher et al. \cite{zssr} pointed out that the visual entropy within a single image is significantly lower than that of a general external image collection, making it feasible to train neural networks using a single image.
Shocher et al. \cite{zssr} were among the first to introduce the concept of internal learning into image super-resolution, demonstrating that a model can learn task-specific knowledge directly from the test image itself. Following this line of research, methods such as PsDip \cite{psdip} and ZS-Pan \cite{zspan} have been proposed. These methods are trained only at test time, effectively alleviating the train-test distribution mismatch, and have demonstrated promising performance on real-world  pansharpening datasets.

However, traditional zero-shot pansharpening methods typically require training a separate set of model parameters for each individual image pair, leading to substantial computational and hardware overhead. In remote sensing imagery, visual structures and textures are often highly similar across different scenes \cite{wang2023remote}, which suggests that it may be possible to design a zero-shot pansharpening framework whose learned weights can generalize across different image pairs.

\subsection{Implicit Neural Representation}
Implicit Neural Representation (INR) \cite{INR1,INR2,INR3} is an emerging framework for continuous signal modeling. Unlike traditional explicit representations, such as pixel arrays or grids, INR employs a multi-layer perceptron (MLP) as a function approximator to directly map spatial coordinates (optionally combined with deep features) to target signal values. This formulation enables end-to-end modeling of complex signals without requiring predefined analytical expressions.

Recent feature-based INR methods no longer learn an independent implicit representation for each object; instead, they share a common function space across different objects. A representative example is LIIF \cite{liif}, which achieves impressive performance in continuous image super-resolution at arbitrary resolutions. To address aliasing artifacts at extremely small scaling factors. UltraSR \cite{ultrasr} further enhances representation capacity by incorporating residual structures and repeatedly injecting coordinate information during the two-dimensional INR process. In the image fusion domain, methods such as NeSSR \cite{nessr} and INF3 \cite{infff} have also demonstrated strong performance, achieving both improved reconstruction quality and the ability to generalize a single set of parameters across multiple objects.

In this work, we adopt a feature-enhanced INR-based fusion network as the core model of the proposed G-ZAP framework, where a feature-based INR, unlike coordinate-based INR that directly maps coordinates to signals \cite{clorf}, enables weight sharing for multi-scale collaborative learning under a semi-supervised training paradigm, leading to high-fidelity reconstruction and great generalization.

\section{Method}
\begin{figure}[t]
    \centering
    \includegraphics[width=0.95\linewidth]{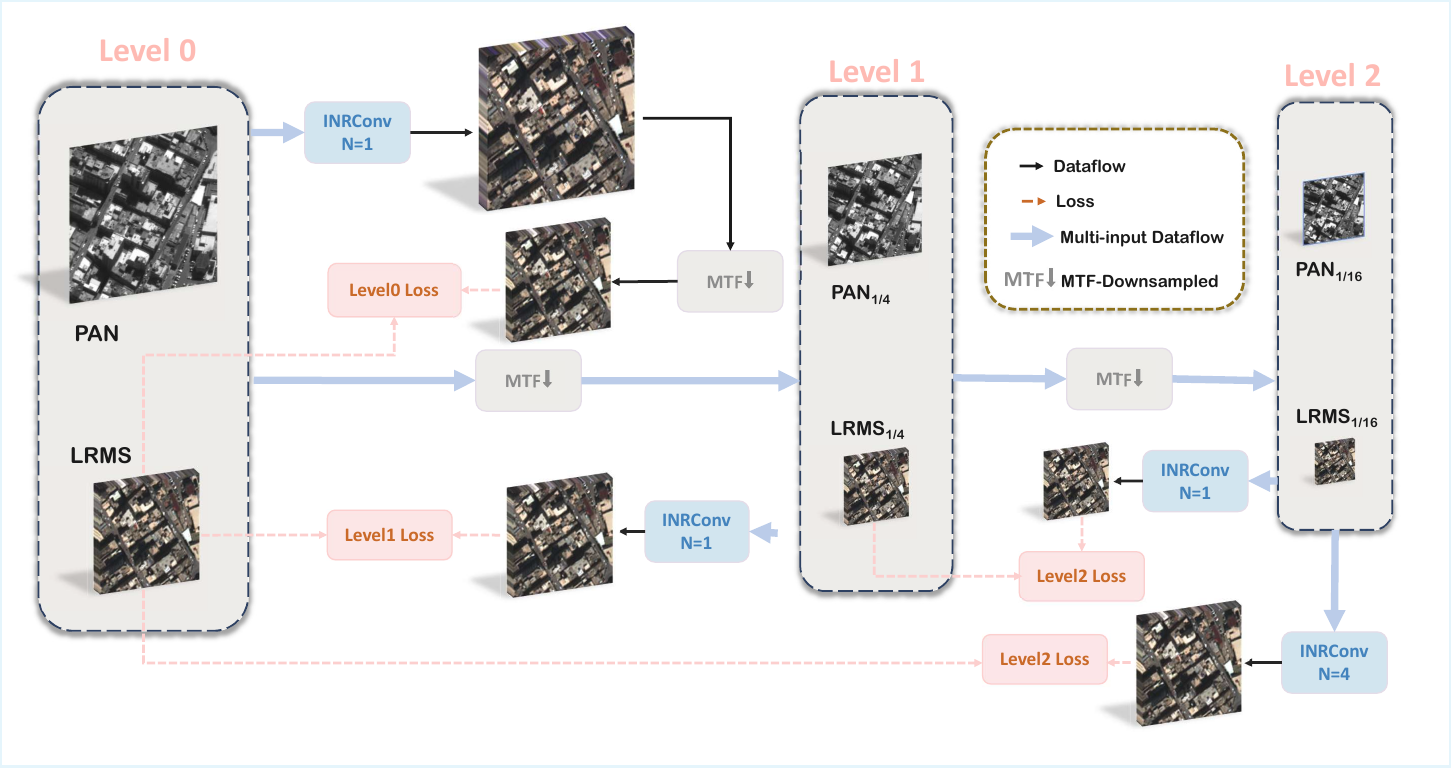}
    \caption{The framework adopts a three-level collaborative training strategy. All INRConv modules share the same parameters across different levels. By jointly optimizing constructed supervisory losses across multiple cross-resolution settings and unsupervised losses at full resolution, the model learns resolution-consistent and generalizable representations. After training, the learned INR-based fusion model can be seamlessly extended to inference at arbitrary spatial resolutions.}
    \label{fig:flowchart}
\end{figure}
In this section, we first introduce the notations used throughout the paper. We then present the overall framework of the proposed G-ZAP method. Subsequently, we provide detailed descriptions of the INRConv model as well as the design of the three-level training strategy, including the formulations and loss functions of Level 0, Level 1, and Level 2.

\subsection{Notations}
The notations used in this paper are defined as follows. Let $\mathbf{P} \in \mathbb{R}^{H \times W}$ denote the high-resolution panchromatic (PAN) image and $\mathcal{Y} \in \mathbb{R}^{h \times w \times c}$ denote the low-resolution multispectral (LRMS) image, where $H,W$ and $h,w$ represent the spatial dimensions and $c$ is the number of spectral bands. We define $r = \tfrac{H}{h} = \tfrac{W}{w}$ as the resolution ratio between PAN and LRMS images.
Let $N$ denote the super-resolution scale factor. The target high-resolution multispectral image at scale $N$ is denoted by $\mathcal{X}_{\times N} \in \mathbb{R}^{(H \times N) \times (W \times N) \times c}$. For multi-level training, the downsampled PAN and LRMS images are represented as $\mathbf{P}_{1/r}$, $\mathbf{P}_{1/r^2}$ and $\mathcal{Y}_{1/r}$, $\mathcal{Y}_{1/r^2}$, respectively.

\subsection{Overall framework}
The flowchart of the proposed G-ZAP framework is shown in \cref{fig:flowchart}. The framework is composed of an INR-based pansharpening backbone, together with a three-level collaborative training strategy. These three levels are designed to preserve full-resolution fidelity, strengthen fusion capability via reduced-resolution supervision, and enable multi-scale generalization, respectively. In the following, we first introduce the INRConv model and then detail the design and loss formulation of each training level.

\subsection{INRConv model}

\begin{figure}[t]
    \centering
    \includegraphics[width=0.99\linewidth]{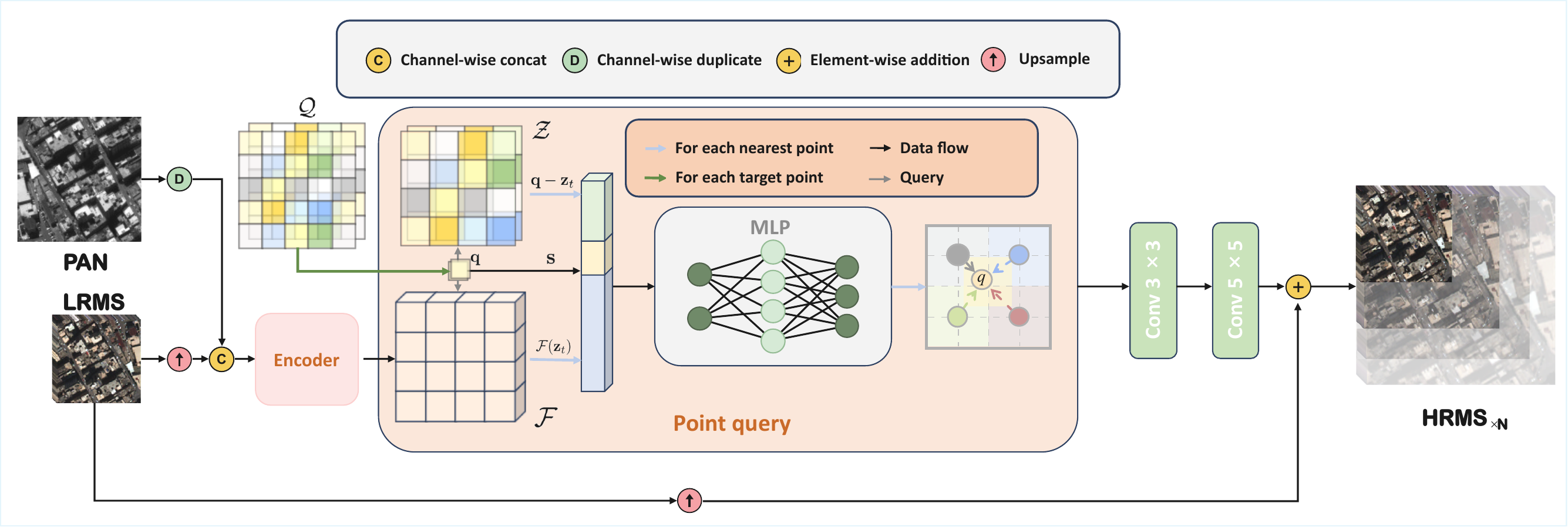}
    \caption{Target coordinates are generated according to the specified scale factor. For each target point, the MLP takes relative coordinates, cell size, and neighboring feature-map values as input, and computes the query value via area-weighted aggregation. The resulting target image is then refined by two lightweight convolutional layers to produce the final output.}
    \label{fig:INR_conv}
\end{figure}

As discussed in \cite{infff}, inserting lightweight convolution layers after an INR network is beneficial for alleviating overfitting. Following this insight, we adopt an INR-based architecture as illustrated in \cref{fig:INR_conv}. The proposed INRConv backbone consists of three components: an encoder for feature extraction, an INR-based point query module for continuous representation, and lightweight convolution layers for decoding. Overall, the proposed model performs inference by taking the PAN image $\mathbf{P}$, the LRMS image $\mathcal{Y}$, and the target super-resolution scale $N$ as inputs, and generates the corresponding HRMS image $\mathcal{X}_{\times N}$:
\begin{equation}
\mathcal{X}_{\times N} = \mathcal{N}_\theta(\mathbf{P}, \mathcal{Y}, N),
\quad \theta=\theta_1 \cup \theta_2,
\end{equation}
where $\mathcal{N}$ denotes the proposed INR-based fusion function and $\theta_1,\theta_2$ are the parameters of the submodules.
Given a PAN image $\mathbf{P}$ and an LRMS image $\mathcal{Y}$, we first duplicate $\mathbf{P}$ along the channel dimension to match the spectral bands of $\mathcal{Y}$, resulting in $\mathcal{P}^D$. Meanwhile, $\mathcal{Y}$ is upsampled to the PAN resolution with scale ratio $r=\tfrac{H}{h}$, denoted as $\widehat{\mathcal{Y}}_{\times r}$. The concatenation of $\mathcal{P}^D$ and $\widehat{\mathcal{Y}}_{\times r}$ is then fed into an EDSR-based \cite{edsr} encoder $E(\cdot)$ to extract a fused feature map
\begin{equation}
\mathcal{F} = E\!\left(\mathcal{P}^D, \widehat{\mathcal{Y}}_{\times r}\right),
\label{eq:feature_map}
\end{equation}
where $\mathcal{F}\in \mathbb{R}^{H \times W \times D}$ is defined on the discrete PAN grid and has $D$ channels.
Given a target super-resolution scale $N$, we define the target coordinate grid $\mathcal{Q}$ on the normalized domain $[-1,1]^2$, where each coordinate is generated as
\begin{equation}
\mathcal{Q}(i,j)=
\left[
-1+\frac{2i+1}{HN},\;
-1+\frac{2j+1}{WN}
\right],
\label{eq:coord}
\end{equation}
with $i\in[0,HN-1]$ and $j\in[0,WN-1]$.
Similarly, the feature coordinate grid $\mathcal{Z}$ is defined on the discrete spatial domain of $\mathcal{F}$ without scale expansion. Given a target point $\mathbf{q}\in\mathcal{Q}$, we perform a point query by identifying its four nearest neighboring feature points in $\mathcal{Z}$, denoted as $\mathbf{z}_t$ with $t\in\{00,01,10,11\}$. Here, the indices $\{00,01,10,11\}$ correspond to the top-left, top-right, bottom-left, and bottom-right neighbors of $\mathbf{q}$, respectively.

For each neighboring point $\mathbf{z}_t$, the local feature $\mathcal{F}(\mathbf{z}_t)$, the relative coordinate $\mathbf{q}-\mathbf{z}_t$, and the cell size $\mathbf{s}$ (encoding the target scale $N$) are directly fed into an MLP $f_{\Theta_1}$ to predict a latent response:

\begin{equation}
\mathbf{v}_t = f_{\Theta_1}\!\left(\mathcal{F}(\mathbf{z}_t),\, \mathbf{q}-\mathbf{z}_t,\, \mathbf{s}\right).
\label{eq:mlp}
\end{equation}                                                                             
The queried feature at $\mathbf{q}$ is obtained by area-weighted aggregation, following LIIF-style \cite{liif} interpolation:
\begin{equation}
\mathbf{v}_{\mathbf{q}} =
\sum_{t\in\{00,01,10,11\}} \frac{S_t}{S} \cdot \mathbf{v}_t,
\quad S=\sum_t S_t,
\label{eq:aggregation}
\end{equation}
where $S_t$ denotes the diagonal sub-area associated with $\mathbf{z}_t$.
By querying all target coordinates in parallel, we obtain a dense feature map $\mathcal{F}_{\times N}$ at the desired resolution. Finally, two lightweight convolution layers, denoted as $g(\cdot)$ with parameters $\theta_2$, decode the queried features to the target HRMS image:
\begin{equation}
\mathcal{X}_{\times N} = g_{\theta_2}\!\left(\mathcal{F}_{\times N}\right).
\label{eq:decode}
\end{equation}

\subsection{Level 0: Full-Resolution Spectral Consistency}

To fully exploit information at full resolution and maintain consistency between training and testing distributions, we introduce an unsupervised loss at the original PAN resolution. In the context of pansharpening, constructing unsupervised spatial losses often requires complex modeling and incurs additional computational overhead. Instead, we adopt a more interpretable and efficient unsupervised spectral consistency loss.

Specifically, under the setting of $N=1$, the INRConv model directly produces an HRMS estimate $\mathcal{N}_{\theta}(\mathbf{P}, \mathcal{Y}, 1)$ at the original PAN resolution. To enforce spectral consistency, the generated HRMS image is first degraded by a sensor-specific modulation transfer function (MTF) \cite{mtf1,mtf2} and then downsampled to the LRMS resolution. The resulting image is required to be consistent with the observed LRMS input $\mathcal{Y}$. The Level 0 loss is formulated as
\begin{equation}
\mathcal{L}_{(0)}
=
\left\|
\mathcal{Y}
-
\downarrow \!\left(
\mathtt{MTF}\!\left(
\mathcal{N}_{\theta}(\mathbf{P}, \mathcal{Y}, 1)
\right)
\right)
\right\|_1 ,
\label{eq:level0}
\end{equation}
where $\mathtt{MTF}(\cdot)$ denotes the sensor-dependent spatial blurring operation, $\downarrow(\cdot)$ indicates downsampling to the LRMS resolution, and $\|\cdot\|_1$ is the $l_1$ norm.
This loss encourages the generated HRMS image to preserve spectral fidelity at full resolution without relying on any ground-truth supervision, thereby aligning the training objective with real-world inference conditions.

\subsection{Level 1: Reduced-Resolution Constructed Supervision}

While the unsupervised spectral consistency at full resolution ensures training-testing distribution alignment, it alone is insufficient to fully enhance the spatial-spectral fusion capability of the model. To this end, we introduce a reduced-resolution training level, where constructed supervision and unsupervised constraints are jointly exploited to further strengthen the representation capacity of the INRConv model.

Specifically, we first blur both the PAN image $\mathbf{P}$ and the LRMS image $\mathcal{Y}$ with a sensor-specific MTF and then downsample them by a factor of $r$, yielding $\mathbf{P}_{1/r}$ and $\mathcal{Y}_{1/r}$, respectively. These degraded observations are then fed into the INRConv model under the setting of $N=1$ to generate a reconstructed multispectral image at the reduced resolution. Since the original LRMS image $\mathcal{Y}$ serves as a valid reference at this scale, we impose a constructed supervision loss by minimizing the $\ell_1$ distance between the model output and $\mathcal{Y}$:
\begin{equation}
\mathcal{L}_{(1)}
=
\left\|
\mathcal{N}_{\theta}(\mathbf{P}_{1/r}, \mathcal{Y}_{1/r}, 1)
-
\mathcal{Y}
\right\|_1 .
\label{eq:level1}
\end{equation}
By combining this constructed supervision with the unsupervised loss at Level 0, the model is encouraged to learn more effective spatial-spectral fusion behavior while preserving consistency with real-world inference conditions.

\subsection{Level 2: Multi-Scale Constructed Supervision}

To further enhance the sensitivity of the model to cell size and improve the multi-scale generalization capability of the INR representation, we introduce a third training level that explicitly supervises the model across different scale factors. Building upon the reduced-resolution setting in Level 1, we further apply sensor-specific MTF blurring followed by downsampling with factor $r$ to construct the Level 2 inputs.

Specifically, the degraded PAN and LRMS images from Level 1 are processed again to obtain $\mathbf{P}_{1/r^2}$ and $\mathcal{Y}_{1/r^2}$. These inputs are fed into the INRConv model with different scale settings to enforce cross-scale consistency. On the one hand, under the setting of $N=1$, the model output is supervised by the reduced-resolution LRMS image $\mathcal{Y}_{1/r}$. On the other hand, under the setting of $N=4$, the model is required to directly reconstruct the original LRMS image $\mathcal{Y}$. The Level 2 loss is thus formulated as
\begin{equation}
\mathcal{L}_{(2)}
=
\left\|
\mathcal{N}_{\theta}(\mathbf{P}_{1/r^2}, \mathcal{Y}_{1/r^2}, 1)
-
\mathcal{Y}_{1/r}
\right\|_1
+
\left\|
\mathcal{N}_{\theta}(\mathbf{P}_{1/r^2}, \mathcal{Y}_{1/r^2}, 4)
-
\mathcal{Y}
\right\|_1 .
\label{eq:level2}
\end{equation}
By explicitly supervising the model at multiple scale factors, this level encourages the INRConv backbone to learn scale-aware representations that generalize across different resolutions.
Finally, the overall training objective is defined as a weighted combination of the losses from the three training levels:
\begin{equation}
\mathcal{L}_{\text{total}}
=
\alpha\,\mathcal{L}_{(0)}
+
\beta\,\mathcal{L}_{(1)}
+
\gamma\,\mathcal{L}_{(2)},
\label{eq:total_loss}
\end{equation}
where $\alpha$, $\beta$, and $\gamma$ are weighting coefficients.

\section{Experiments}
\subsection{Datasets, Metrics, and Implementation Details}
\subsubsection{Datasets and Metrics}

Experiments are conducted on three datasets from PanCollection\footnote{\url{https://github.com/liangjiandeng/PanCollection}}, acquired by the WorldView-3 (WV3), GaoFen-2 (GF2), and WorldView-2 (WV2) satellites. Each dataset provides registered PAN/LRMS pairs, where PAN images are cropped to $512 \times 512$ pixels. WV3 and WV2 contain 8 spectral bands, while GF2 contains 4.

Since our method targets real-world pansharpening without ground-truth references, evaluation is mainly performed at full resolution using three widely adopted no-reference metrics: HQNR~\cite{aiazzi2014full}, $D_s$, and $D_\lambda$. HQNR combines $D_s$ and $D_\lambda$ to jointly assess spatial detail preservation and spectral fidelity.

\subsubsection{Implementation Details}

Only the WV3 and GF2 training sets are used for the pretrained baseline models, corresponding to the 8-band and 4-band configurations, respectively.
In contrast, G-ZAP trains from the test PAN/LRMS pair only and uses no external HR/LR pairs. Evaluation is conducted on the in-domain datasets (WV3 and GF2) as well as the cross-sensor dataset (WV2) to assess generalization across different sensors.
Our INRConv model is optimized using Adam with an initial learning rate of $5 \times 10^{-4}$ for 500 epochs. The loss weights $\alpha, \beta, \gamma$ are set to $1, 1, 0.2$ for the 8-band setting and $1, 1, 4$ for the 4-band setting.
All experiments are conducted on a workstation equipped with an NVIDIA RTX 3090 GPU (24\,GB memory) and an AMD EPYC 7402 24-Core CPU.
\subsection{Benchmarks}
We compare G-ZAP with several representative SOTA pansharpening approaches, including traditional, supervised deep learning (DL), and zero-shot methods.
\textbf{Traditional methods:}  C-BDSD\cite{c-bdsd}, MF\cite{mf}, BT-H\cite{bt-h}, MTF-GLP-FS\cite{mtf-glp-fs}, and BDSD-PC\cite{bdsd-pc}.
\textbf{Pretrained DL methods:} FusionNet\cite{fusionnet}, LAGNet\cite{lagconv}, FusionMamba\cite{fusionmamba}, and WFANet\cite{wfanet}.
\textbf{Zero-shot methods:} PsDip\cite{psdip}, ZS-Pan\cite{zspan}, and UCL \cite{ucl}.

\begin{table}[t]
\centering
\caption{Performance comparison on three real-world datasets: 
GF2, WV3 and WV2. 
The reported values represent the average over 20 test images. 
\textbf{Bold}: best; \underline{Underline}: second best.}
\label{tab:gf2_wv3_wv2_results}

\begin{tabular}{c|ccc|ccc|ccc}
\toprule
\multirow{2}{*}{Method}
& \multicolumn{3}{c|}{GF2}
& \multicolumn{3}{c|}{WV3}
& \multicolumn{3}{c}{WV2} \\
\cmidrule(lr){2-4}\cmidrule(lr){5-7}\cmidrule(lr){8-10}
& HQNR$\uparrow$ & $D_\lambda\downarrow$ & $D_s\downarrow$
& HQNR$\uparrow$ & $D_\lambda\downarrow$ & $D_s\downarrow$
& HQNR$\uparrow$ & $D_\lambda\downarrow$ & $D_s\downarrow$ \\
\midrule

BT-H 
& 0.8923 & 0.0832 & 0.0814
& 0.8659 & 0.0656 & 0.0742
& 0.8300 & 0.0860 & 0.0925 \\

C-BDSD 
& 0.8870 & 0.0619 & 0.0689
& 0.8562 & 0.0874 & 0.0618
& 0.6956 & 0.2253 & 0.1019 \\

BDSD-PC
& 0.8710 & 0.0633 & 0.0742
& 0.8673 & 0.0634 & 0.0749
& 0.8286 & 0.1413 & 0.0356 \\

MTF-GLP-FS
& 0.7423 & 0.1438 & 0.1331
& 0.9127 & 0.0357 & 0.0537
& 0.8658 & 0.0563 & 0.0830 \\

MF
& 0.8730 & 0.0646 & 0.0633
& 0.9014 & 0.0452 & 0.0561
& 0.8508 & 0.0704 & 0.0857 \\

PsDip
& 0.8833 & 0.0448 & 0.0752
& 0.9215 & 0.0191 & 0.0607
& 0.8980 & {0.0385} & 0.0659 \\

ZS-Pan
& 0.8961 & 0.0361 & 0.0705
& 0.9449 & 0.0254 & 0.0306
& 0.9112 & 0.0476 & 0.0435 \\

WFANet
& 0.9463 & \underline{0.0162} & 0.0381
& 0.9437 & \textbf{0.0172} & 0.0390
& 0.9128 & 0.0526 & \textbf{0.0366} \\

FusionNet
& 0.8673 & 0.0350 & 0.1013
& 0.9228 & 0.0424 & 0.0364
& 0.8881 & 0.0543 & 0.0606 \\

LAGNet
& 0.8815 & 0.0423 & 0.0795
& 0.9133 & 0.0389 & 0.0500
& 0.6883 & 0.1696 & 0.1692 \\

UCL
& 0.9427 & \textbf{0.0104} & 0.0474
& 0.9482 & \underline{0.0182} & 0.0344
& 0.9201 & \underline{0.0228} & 0.0585\\

FusionMamba
& {0.9536} & {0.0174} & {0.0295}
& {0.9550} & {0.0183} & 0.0272
& 0.9064 & 0.0526 & \underline{0.0432} \\

\midrule

G-ZAP$^{*}$
& \underline{0.9569} & 0.0188 & \underline{0.0246}
& \underline{0.9574} & 0.0187 & \underline{0.0244}
& \textbf{0.9367} & \textbf{0.0195} & {0.0450} \\
G-ZAP
& \textbf{0.9706} & {0.0181} & \textbf{0.0115}
& \textbf{0.9585} & {0.0187} & \textbf{0.0233}
& \underline{0.9226} & {0.0304} & {0.0486} \\

\bottomrule
\end{tabular}
\end{table}

\subsection{Comparison with State-of-the-Art Methods}

To benchmark our approach against existing methods, we evaluate our framework with a scaling factor of $N=1$ on both the in-domain datasets (WV3 and GF2) and the cross-sensor dataset (WV2). 

To further demonstrate the strong weight reusability of our framework, we additionally report a weight-reuse variant denoted as G-ZAP$^{*}$. Specifically, for the 8-band setting, the model trained on WV3 (id=0) is directly reused for inference on both WV3 and WV2 without any retraining; for the 4-band setting, the weights trained on GF2 (id=0) are directly applied for all testing image pairs. 
\begin{figure}[ht]
    \centering
    \includegraphics[width=1\linewidth]{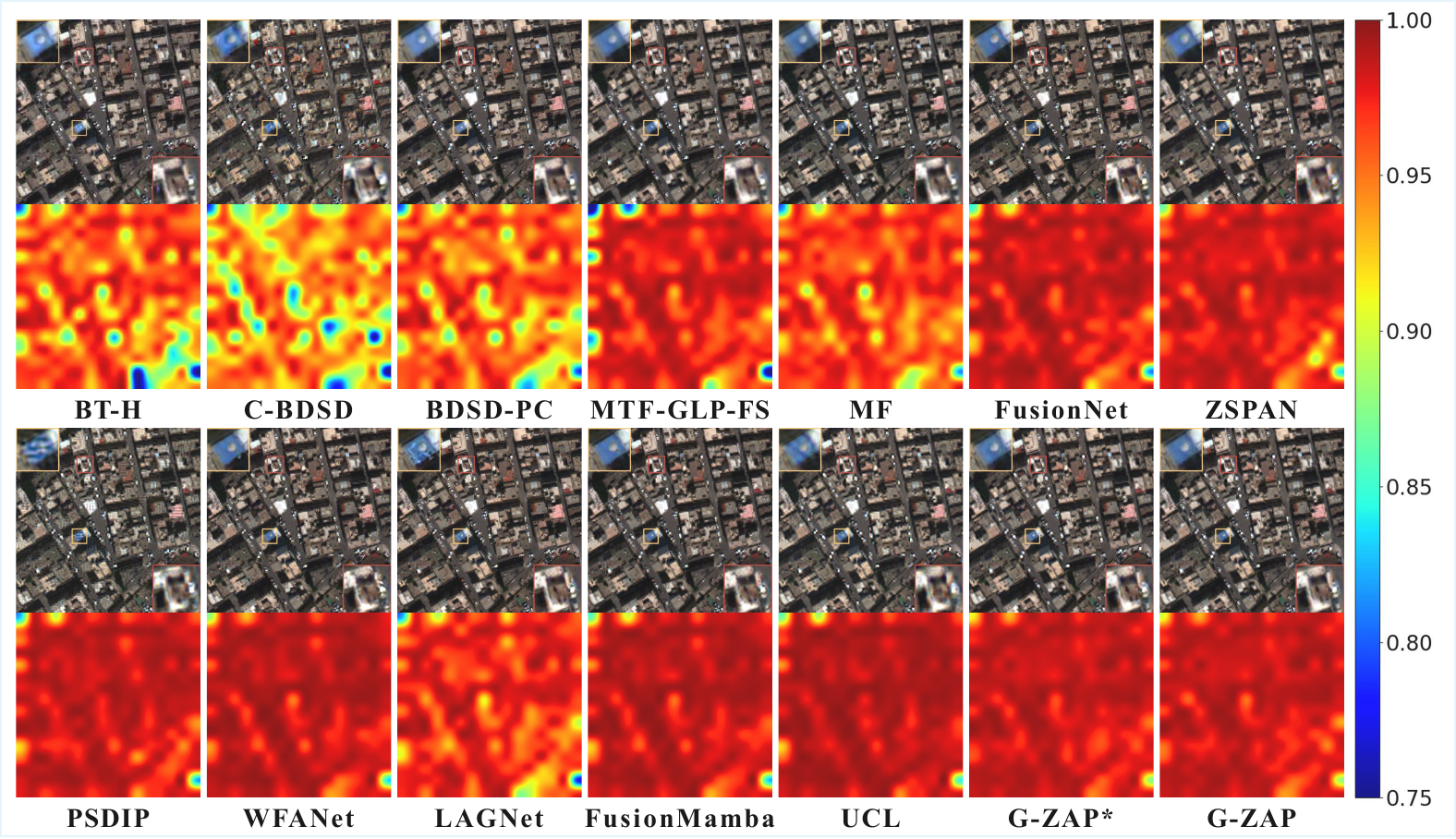}
    \caption{Visual Fusion Image and HQNR Map on a full resolution WV3 example}
    \label{fig:wv3}
\end{figure}
\begin{figure}[ht]
    \centering
    \includegraphics[width=1\linewidth]{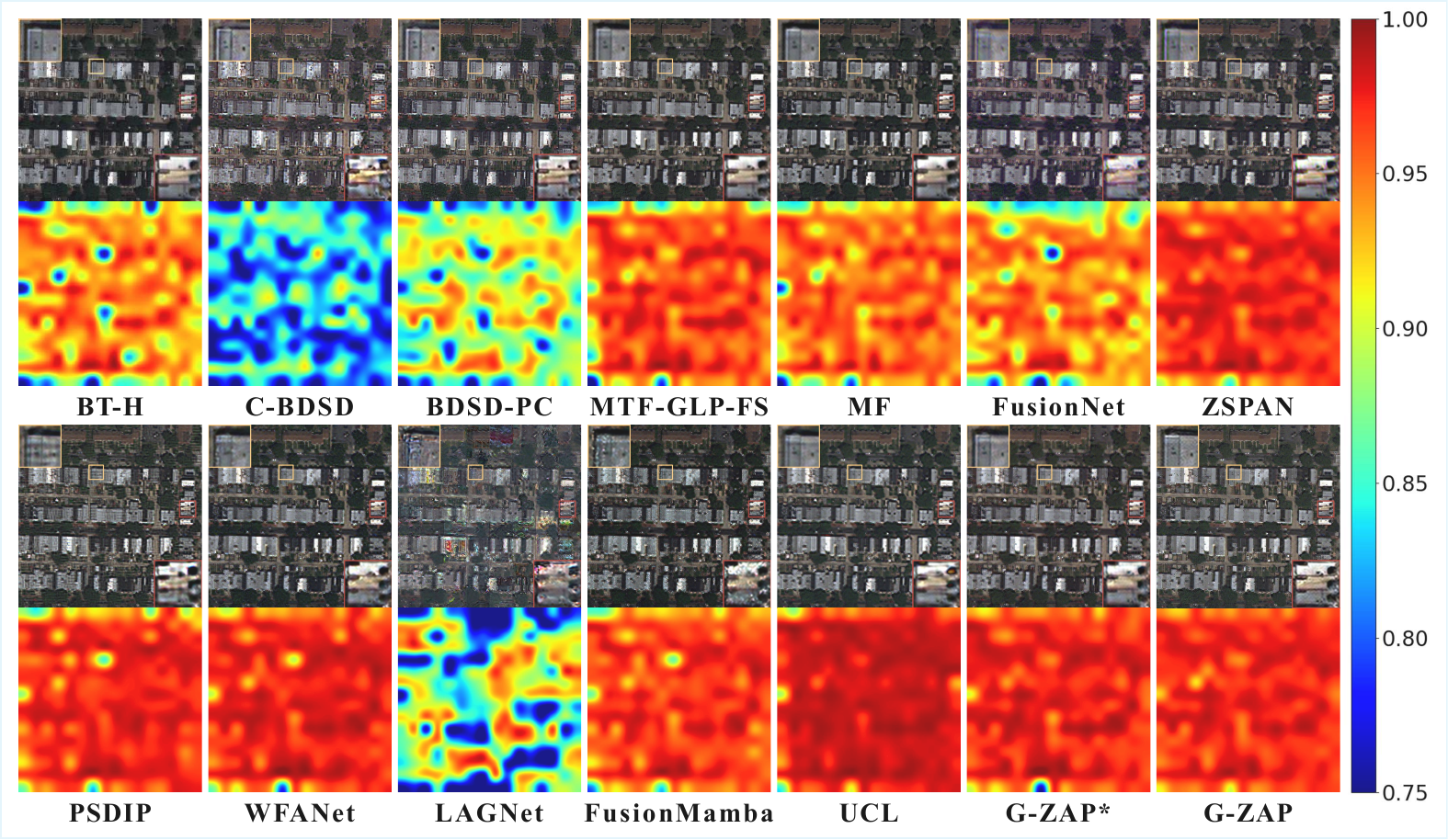}
    \caption{Visual Fusion Image and HQNR Map on a full resolution WV2 example}
    \label{fig:wv2}
\end{figure}
Quantitative comparisons are summarized in \cref{tab:gf2_wv3_wv2_results}, while visual results and the corresponding HQNR maps are presented in \cref{fig:wv3,fig:wv2}. As shown, both the standard per-image zero-shot training--inference scheme and the weight-reuse strategy consistently achieve state-of-the-art performance on in-domain and cross-sensor datasets without spatial fidelity used in other zero-shot method, while producing superior visual quality. These results indicate that, although our method is designed for cross-resolution generalization, it remains highly competitive and achieves the best performance under conventional HRMS resolutions.

\subsection{Arbitrary-Scale Pansharpening}
\begin{figure}[ht]
    \centering
    \includegraphics[width=0.95\linewidth]{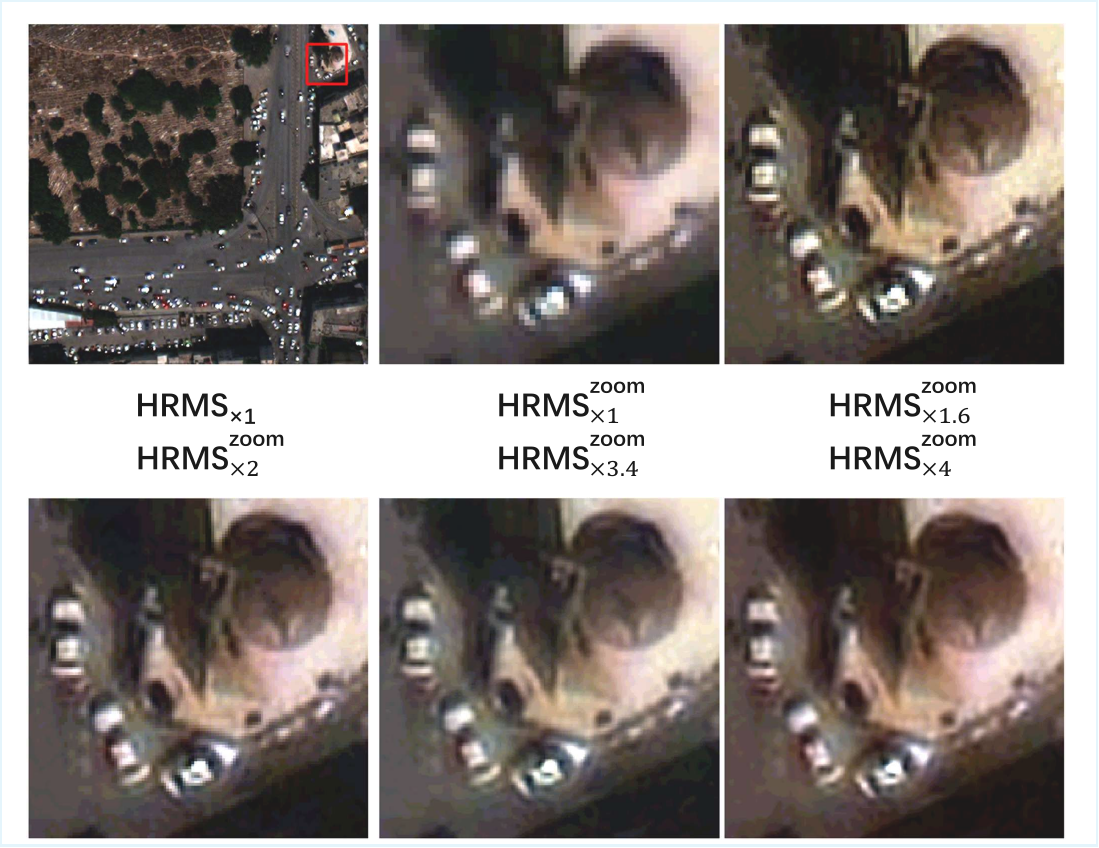}
    \caption{Visual results of arbitrary-scale pansharpening at different magnification ratios ($\times 1$, $\times 1.6$, $\times 2$, $\times 3.4$, and $\times 4$). Zoomed-in regions show high-quality spatial details and spectral fidelity across scales.}
    \label{fig:arbitrary-scale}
\end{figure}
We further evaluate the proposed method under multiple upsampling factors to examine its performance across different output resolutions. Given the same LRMS/PAN input, starting from the standard pansharpened result ($\times 1$), HRMS images are generated at several magnification ratios, including $\times 1$, $\times 1.6$, $\times 2$, $\times 3.4$, and $\times 4$.

As shown in Fig.~\ref{fig:arbitrary-scale}, the proposed method produces visually high-quality results at different scales. Fine structures such as edges, vehicles, and building boundaries remain clear, while spectral characteristics are well preserved. The results demonstrate that the method maintains stable reconstruction quality at different spatial resolutions and provides richer details at high resolutions.
\subsection{Ablation Study}
\begin{table}[t]
\centering
\caption{Average results of the ablation study for our G-ZAP framework on WV3 and WV2 datasets.
\textbf{Bold}: best; \underline{Underline}: second best.}
\label{tab:ablation}

\setlength{\tabcolsep}{3pt}
\renewcommand{\arraystretch}{1.1}

\begin{tabular}{l|ccc|ccc}
\toprule
\multirow{2}{*}{Name}
& \multicolumn{3}{c|}{WV3}
& \multicolumn{3}{c}{WV2} \\
\cmidrule(lr){2-4}\cmidrule(lr){5-7}
& HQNR$\uparrow$ & $D_\lambda\downarrow$ & $D_s\downarrow$
& HQNR$\uparrow$ & $D_\lambda\downarrow$ & $D_s\downarrow$ \\
\midrule

G-ZAP
& \textbf{0.9585} & {0.0187} & \textbf{0.0233}
& \textbf{0.9226} & \underline{0.0304} & \textbf{0.0486} \\

w/o $\mathcal{L}_{(0)}$
& 0.9276 & 0.0279 & 0.0459
& 0.8781 & 0.0496 & 0.0760 \\

w/o $\mathcal{L}_{(1)}$
& {0.9377} & \textbf{0.0181} & {0.0451}
& 0.8970 & 0.0313 & 0.0743 \\

w/o $\mathcal{L}_{(2)}$
& \underline{0.9511} & \underline{0.0185} & \underline{0.0310}
& \underline{0.9178} & \textbf{0.0264} & \underline{0.0574} \\

\bottomrule
\end{tabular}
\end{table}

\subsubsection{The loss used in our framework}
\begin{figure}[ht]
    \centering
    \includegraphics[width=\linewidth]{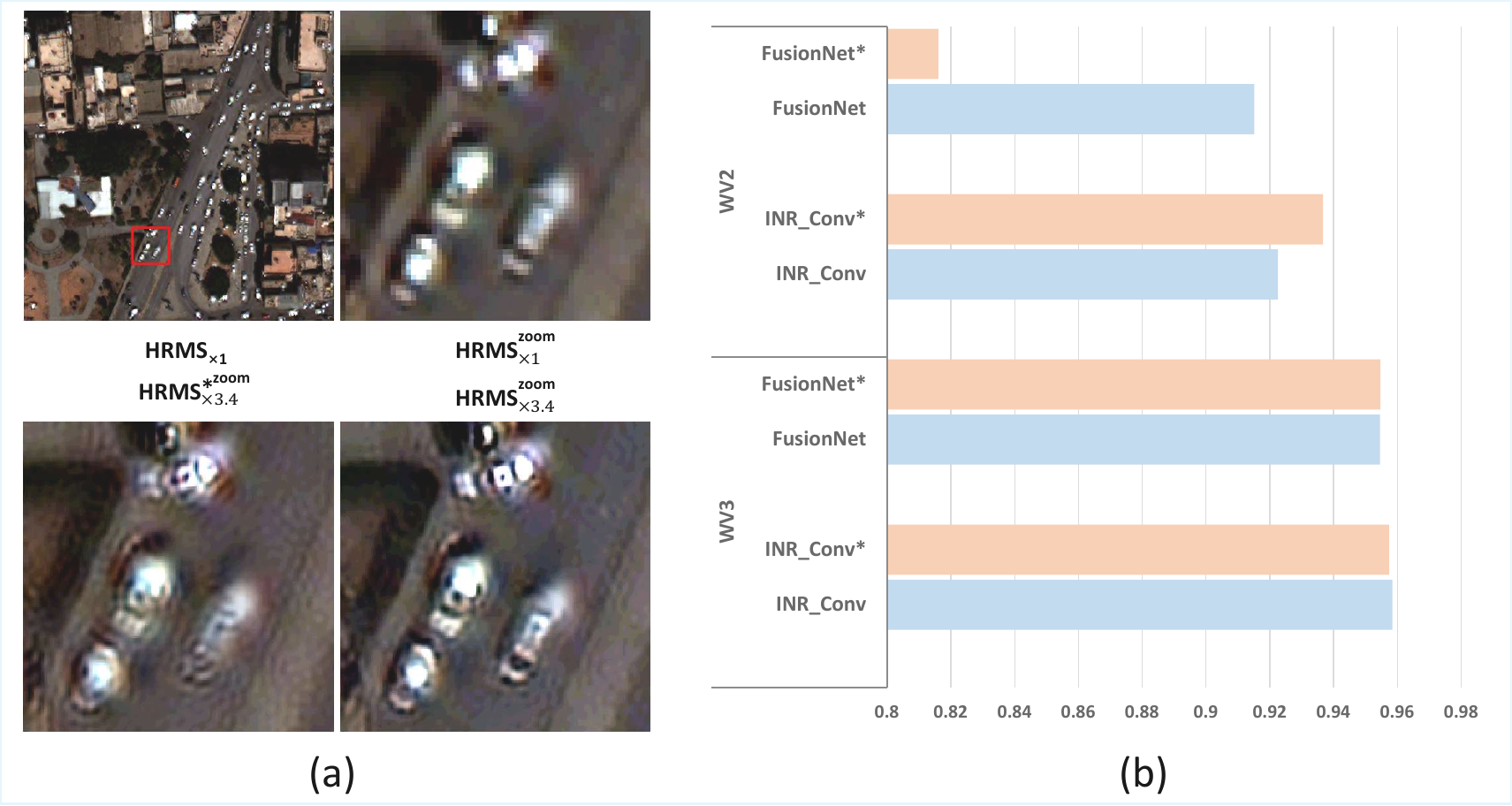}
    \caption{Ablation results. (a) Removing the Level-2 loss $\mathcal{L}_{(2)}$ causes blurred textures and missing fine details ($^{*}$ indicates results without $\mathcal{L}_{(2)}$). (b) Backbone comparison on WV3 and WV2 showing that INRConv outperforms FusionNet ($^{*}$ denotes weight reuse from the WV3 (id=0) model).}
    \label{fig:ablation}
\end{figure}

We conduct ablation experiments to analyze the contribution of each loss component in the proposed framework. Starting from the full model (G-ZAP), we remove one loss term at a time, including $\mathcal{L}_{(0)}$, $\mathcal{L}_{(1)}$, and $\mathcal{L}_{(2)}$, while keeping the remaining settings unchanged.
The quantitative results on both WV3 (in-domain) and WV2 (cross-sensor) datasets are reported in Table~\ref{tab:ablation}. The complete model consistently achieves the best or second-best performance across all metrics. Removing any loss term leads to noticeable degradation in HQNR as well as spatial or spectral distortion measures ($D_\lambda$ and $D_s$). These results demonstrate that each component contributes positively and that the three losses complement each other to achieve optimal performance.
Additionally, Fig.~\ref{fig:ablation} (a) provides a visual comparison with and without the Level-2 loss. Removing $\mathcal{L}_{(2)}$ results in blurrier structures and the loss of fine details in high-resolution regions, whereas the full model generates sharper and more faithful reconstructions. This observation further validates the effectiveness of the proposed framework.
\begin{table}[ht]
\centering
\caption{Average performance comparison of zero-shot methods and the proposed G-ZAP framework on WV3 and WV2 datasets.
\textbf{Bold}: best; \underline{Underline}: second best. Running time in seconds.}
\label{tab:time}

\setlength{\tabcolsep}{3pt}
\renewcommand{\arraystretch}{1.1}

\begin{tabular}{l|cccc|cccc}
\toprule
\multirow{2}{*}{Method}
& \multicolumn{4}{c|}{WV3}
& \multicolumn{4}{c}{WV2} \\
\cmidrule(lr){2-5}\cmidrule(lr){6-9}
& HQNR$\uparrow$ & $D_\lambda\downarrow$ & $D_s\downarrow$ & Time$\downarrow$
& HQNR$\uparrow$ & $D_\lambda\downarrow$ & $D_s\downarrow$ & Time$\downarrow$ \\
\midrule

PsDip
& 0.9215 & {0.0191} & 0.0607 & 285.37
& 0.8980 & 0.0385 & 0.0659 & 280.23 \\

ZS-Pan
& 0.9449 & 0.0254 & 0.0306 & 67.24
& 0.9112 & 0.0476 & \textbf{0.0435} & 68.52 \\

UCL
& 0.9482 & \underline{0.0182} & 0.0344& {463.22}
& 0.9201 & \underline{0.0228} & 0.0585& {464.41} \\

G-ZAP
& \textbf{0.9585} & \underline{0.0187} & \underline{0.0233}& \underline{79.03}
& \underline{0.9226} & {0.0304} & {0.0486}& \underline{78.58} \\

G-ZAP$^*$
& \underline{0.9574} & \underline{0.0187} & \textbf{0.0244}& \textbf{4.23}
& \textbf{0.9367} & \textbf{0.0195} & \underline{0.0450} & \textbf{0.14} \\
\bottomrule
\end{tabular}
\end{table}

\subsubsection{INRConv}
To evaluate the effectiveness of the proposed INRConv backbone within the G-ZAP framework, we compare it with a conventional fully convolutional architecture (FusionNet) under identical training and evaluation settings. For a fair comparison, INRConv is replaced with FusionNet while keeping all other components unchanged, and the $N=4$ branch of $\mathcal{L}_2$ is removed accordingly.
Fig.~\ref{fig:ablation} (b) reports the quantitative results on the WV3 and WV2 datasets. INRConv consistently achieves superior performance across different settings. Notably, in the weight-reuse evaluation on WV2, FusionNet exhibits a significant performance drop compared with its zero-shot training counterpart, suggesting limited cross-sensor transferability of purely convolutional features. In contrast, INRConv maintains stable performance, demonstrating stronger generalization capability and better suitability for weight reuse. We attribute this advantage to the feature-based INR formulation, which can be viewed as a more expressive interpolation mechanism. Such a property allows the model to better exploit structural and textural similarities in remote sensing imagery, leading to improved generalization, particularly under cross-sensor scenarios where image characteristics may vary. These results validate the effectiveness of the proposed INRConv in our framework.

\subsection{Running Time}

The runtime comparison is reported in Table~\ref{tab:time}. Benefiting from the weight-reuse strategy, the training cost can be amortized across images, eliminating the need for per-image optimization. As a result, the proposed method achieves extremely fast inference, requiring only a few seconds or even sub-second time per image.
Both in-domain (WV3) and cross-sensor (WV2) evaluations show that G-ZAP$^{*}$ attains the lowest running time while maintaining superior quantitative performance, demonstrating the high efficiency and practical applicability of our framework.

\section{Conclusion}

This paper presents G-ZAP, a general zero-shot framework for pansharpening across arbitrary resolutions. The framework integrates a feature-based INR fusion backbone with a carefully designed  multi-level semi-supervised training strategy, enabling strong cross-resolution, cross-scene, and cross-sensor generalization without requiring large-scale pre-training. In addition, the design naturally supports efficient weight reuse across image pairs, significantly reducing computational cost, requiring only a few seconds or even sub-second per image, thereby making it highly practical and efficient for real-world applications. Extensive experiments on multiple real-world datasets demonstrate that G-ZAP achieves state-of-the-art performance in both quantitative metrics and visual quality.

\section*{Acknowledgements}
This research was supported by The Project of the Department of Science and Technology of Sichuan Province (Grant No. 2025YFNH0001).
%
%
\bibliographystyle{splncs04}
\bibliography{main}

\clearpage
\appendix
\renewcommand{\thesection}{S\arabic{section}}
\setcounter{section}{0}
\section*{Supplementary Material}
\section{Results on Simulated Datasets}

Although our method is primarily designed for real-world arbitrary-scale pansharpening, we also report its performance under specific resolution settings. As shown in Table \ref{tab:a}, although no finely designed spatial loss is introduced,  the proposed method achieves superior quantitative performance compared with all traditional methods and ZS-Pan on both the WV3 and WV2 datasets.

\begin{table}[ht]
\centering
\caption{The average results on simulated WV3, WV2 datasets}
\label{tab:a}
\begin{tabular}{c|cccc|cccc}
\toprule
\multirow{2}{*}{Method}
& \multicolumn{4}{c|}{WV3}
& \multicolumn{4}{c}{WV2} \\
\cmidrule(lr){2-5}\cmidrule(lr){6-9}
& Q2n$\uparrow$ & SAM$\downarrow$ & ERGAS$\downarrow$ & SCC$\uparrow$
& Q2n$\uparrow$ & SAM$\downarrow$ & ERGAS$\downarrow$ & SCC$\uparrow$ \\
\midrule

BT-H
& 0.8192 & \underline{4.8655} & 4.6520 & {0.9261}
& 0.8233 & \underline{5.9818} & 4.5236 & 0.9153 \\

C-BDSD
& 0.7980 & 6.1701 & 6.0917 & 0.8868
& 0.8182 & 7.4277 & 5.0267 & 0.8977 \\

BDSD-PC
& \underline{0.8267} & 5.5398 & 4.7187 & 0.9054
& \underline{0.8457} & 6.3109 & \underline{4.3116} & 0.9088 \\

MTF-GLP-FS
& 0.8191 & 5.4700 & 4.9448 & 0.8952
& 0.8202 & 6.4064 & 4.7065 & 0.8877 \\

MF
& 0.8066 & 5.3907 & 5.1392 & 0.9047
& 0.8152 & 6.3253 & 4.8677 & 0.8966 \\

ZS-Pan
& 0.8118 & 5.3000 & \textbf{4.4397} & \underline{0.9339}
& {0.8441} & {6.1298} & \textbf{4.2345} & \textbf{0.9202} \\

\midrule

G-ZAP$^{*}$
& 0.8129 & 5.4983 & 4.7895 & 0.9160
& 0.8382 & 6.0053 & {4.4211} & {0.9073} \\

G-ZAP
& \textbf{0.8346} & \textbf{4.7577} & \underline{4.4500} & \textbf{0.9344}
& \textbf{0.8461} & \textbf{5.8937} & {4.3500} & \underline{0.9169} \\

\bottomrule
\end{tabular}
\end{table}

\section{Quantitative Arbitrary-Scale Evaluation}

To further verify the arbitrary-scale capability of G-ZAP, we conduct a reference-based quantitative evaluation on reduced-resolution WV3 data, where ground-truth references are available. We evaluate unseen non-standard scales and compare direct arbitrary-scale querying of G-ZAP with two simple baselines: bicubic interpolation and the strategy of first generating the standard $\times1$ G-ZAP output followed by bicubic upsampling, denoted as G-ZAP$_{\times1}$+Bic.

As shown in Table~\ref{tab:arbitrary_scale}, directly querying G-ZAP at the target scale consistently outperforms both baselines on most metrics, especially at larger scales. This demonstrates that G-ZAP does not merely resize the $\times1$ output, but benefits from the scale-conditioned implicit representation to perform scale-aware spatial-spectral reconstruction.

\begin{table}[ht]
\centering
\caption{Quantitative arbitrary-scale results on reduced-resolution WV3 data. Bic. denotes bicubic interpolation. G-ZAP$_{\times1}$+Bic. denotes first generating the standard $\times1$ G-ZAP output and then applying bicubic upsampling to the target scale.}
\label{tab:arbitrary_scale}
\begin{tabular}{c|c|cccc}
\toprule
Scale & Method & Q2n$\uparrow$ & SAM$\downarrow$ & ERGAS$\downarrow$ & SCC$\uparrow$ \\
\midrule
\multirow{3}{*}{2.1}
& Bic. & 0.4217 & 8.22 & 9.10 & 0.6319 \\
& G-ZAP$_{\times1}$+Bic. & 0.5848 & 8.83 & 9.25 & 0.7745 \\
& G-ZAP & \textbf{0.6163} & \textbf{7.18} & \textbf{8.70} & \textbf{0.7968} \\
\midrule
\multirow{3}{*}{3.4}
& Bic. & 0.2331 & 8.99 & 11.40 & 0.5472 \\
& G-ZAP$_{\times1}$+Bic. & 0.5571 & 11.55 & 11.25 & 0.7552 \\
& G-ZAP & \textbf{0.6208} & \textbf{8.22} & \textbf{8.55} & \textbf{0.7774} \\
\midrule
\multirow{3}{*}{6.7}
& Bic. & 0.1534 & 10.09 & 12.55 & 0.4479 \\
& G-ZAP$_{\times1}$+Bic. & 0.4107 & 13.21 & 13.80 & 0.6892 \\
& G-ZAP & \textbf{0.4926} & \textbf{8.73} & \textbf{9.70} & \textbf{0.7135} \\
\bottomrule
\end{tabular}
\end{table}

\section{Additional Full-Resolution Comparison}

To strengthen the comparison with recent methods and feature-based INR fusion models, we further include ADWM \cite{cvpr2025}, Pan-Crafter \cite{iccv2025}, and INF3 \cite{infff} under the same full-resolution evaluation protocol. ADWM and Pan-Crafter are recent conference methods, while INF3 is a representative feature-based INR fusion baseline. As shown in Table~\ref{tab:additional_baselines}, G-ZAP achieves the best HQNR on both WV3 and GF2. Compared with INF3, G-ZAP obtains better overall performance, indicating that the proposed zero-shot training framework is important for adapting feature-based INR representations to real-world pansharpening.

\begin{table}[ht]
\centering
\caption{Additional full-resolution comparison with recent methods and a feature-based INR baseline.}
\label{tab:additional_baselines}
\begin{tabular}{c|ccc|ccc}
\toprule
\multirow{2}{*}{Method}
& \multicolumn{3}{c|}{WV3}
& \multicolumn{3}{c}{GF2} \\
\cmidrule(lr){2-4}\cmidrule(lr){5-7}
& HQNR$\uparrow$ & $D_{\lambda}\downarrow$ & $D_s\downarrow$
& HQNR$\uparrow$ & $D_{\lambda}\downarrow$ & $D_s\downarrow$ \\
\midrule
ADWM & 0.955 & 0.019 & 0.026 & 0.931 & 0.021 & 0.049 \\
Pan-Crafter & 0.958 & \textbf{0.016} & 0.027 & 0.964 & 0.019 & 0.017 \\
INF3 & 0.949 & 0.017 & 0.035 & 0.943 & 0.021 & 0.037 \\
G-ZAP & \textbf{0.959} & 0.019 & \textbf{0.023} & \textbf{0.971} & \textbf{0.018} & \textbf{0.012} \\
\bottomrule
\end{tabular}
\end{table}

\section{More Visual Comparison}
To provide a more intuitive demonstration of the visual performance of our method for arbitrary-scale pansharpening, we include supplementary videos that compare our arbitrary-scale outputs with conventional upsampling methods that first generate the standard HRMS×1 result and then upsample it by nearest-neighbor or bicubic interpolation.

\begin{figure}[ht]
    \centering
    \includegraphics[width=1\linewidth]{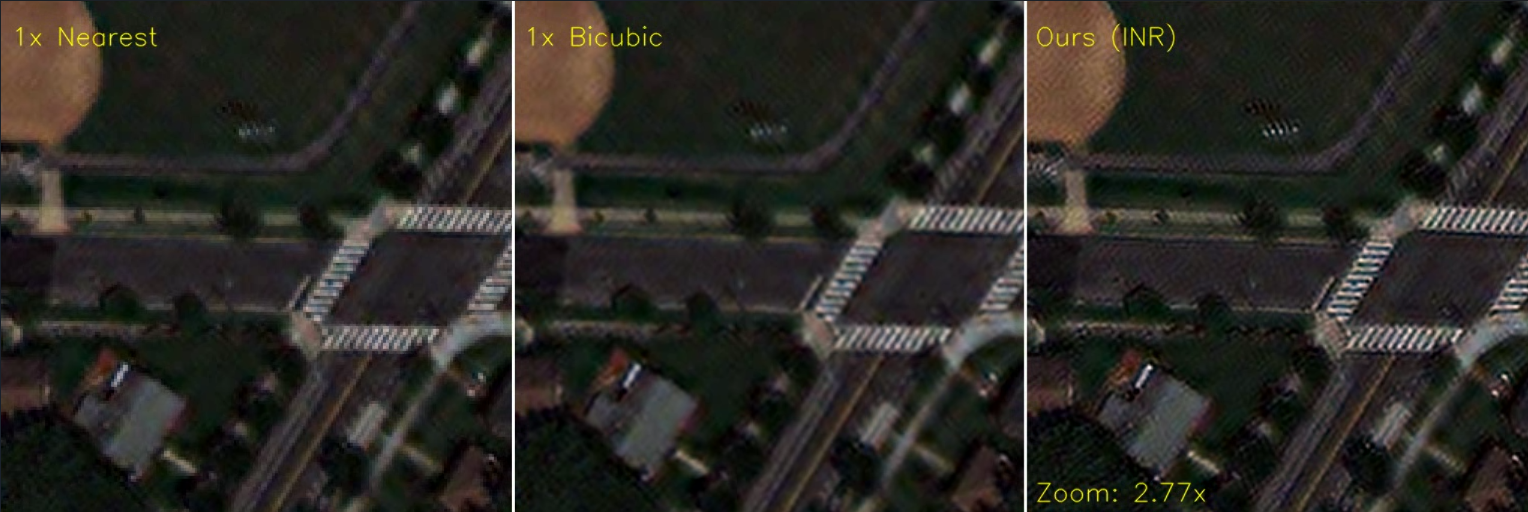}
    \caption{Visual comparison of nearest-neighbor upsampling, bicubic interpolation, and G-ZAP at a scale factor of 2.77 on WV2, where G-ZAP preserves clearer details.}
    \label{fig:a}
\end{figure}

\begin{figure}[ht]
    \centering
    \includegraphics[width=1\linewidth]{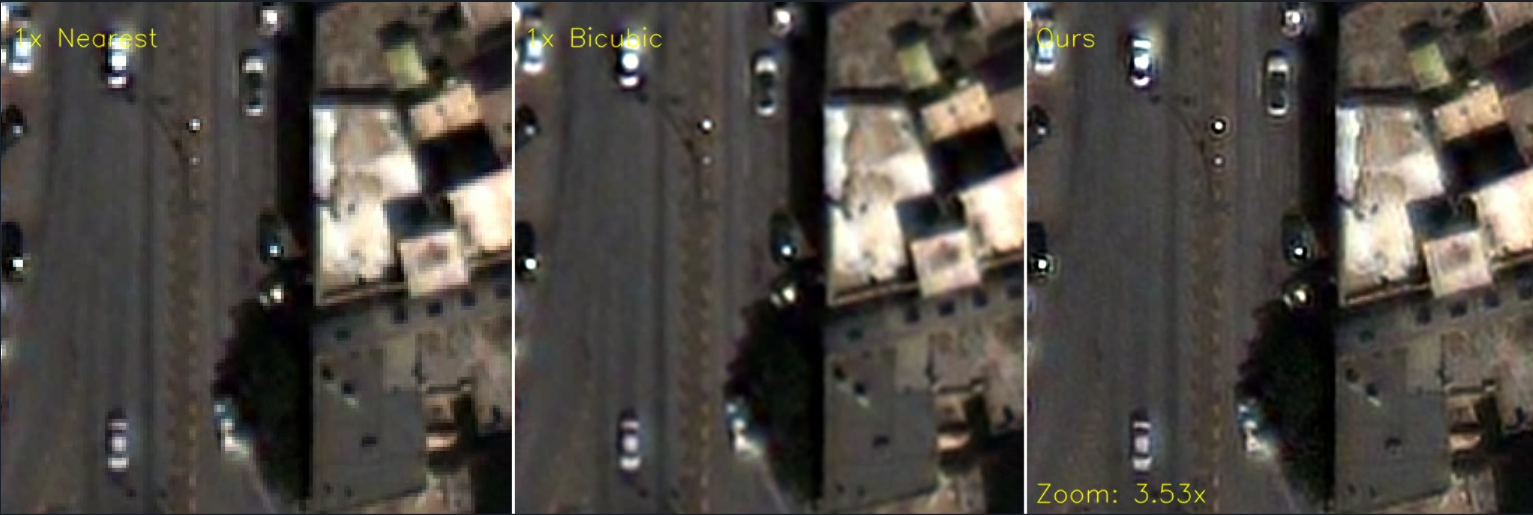}
    \caption{Visual comparison of nearest-neighbor upsampling, bicubic interpolation, and G-ZAP at a scale factor of 3.53 on WV3, where G-ZAP preserves clearer details.}
    \label{fig:a2}
\end{figure}

The videos are provided in the supplementary files and illustrate the visual differences across multiple scenes. Figure \ref{fig:a} and Figure \ref{fig:a2} show a representative frame from the video. As observed, compared with the blurred and over-smoothed results produced by conventional upsampling methods, our method preserves significantly clearer and sharper details.
\section{Loss--Epoch Convergence Curve}

The number of training epochs is a key factor affecting the final performance of the model. In practice, there exists a trade-off between the number of epochs and the degree of convergence. Using fewer epochs can reduce the training cost but may lead to insufficient convergence and consequently degraded performance.

\begin{figure}[ht]
    \centering
    \includegraphics[width=1\linewidth]{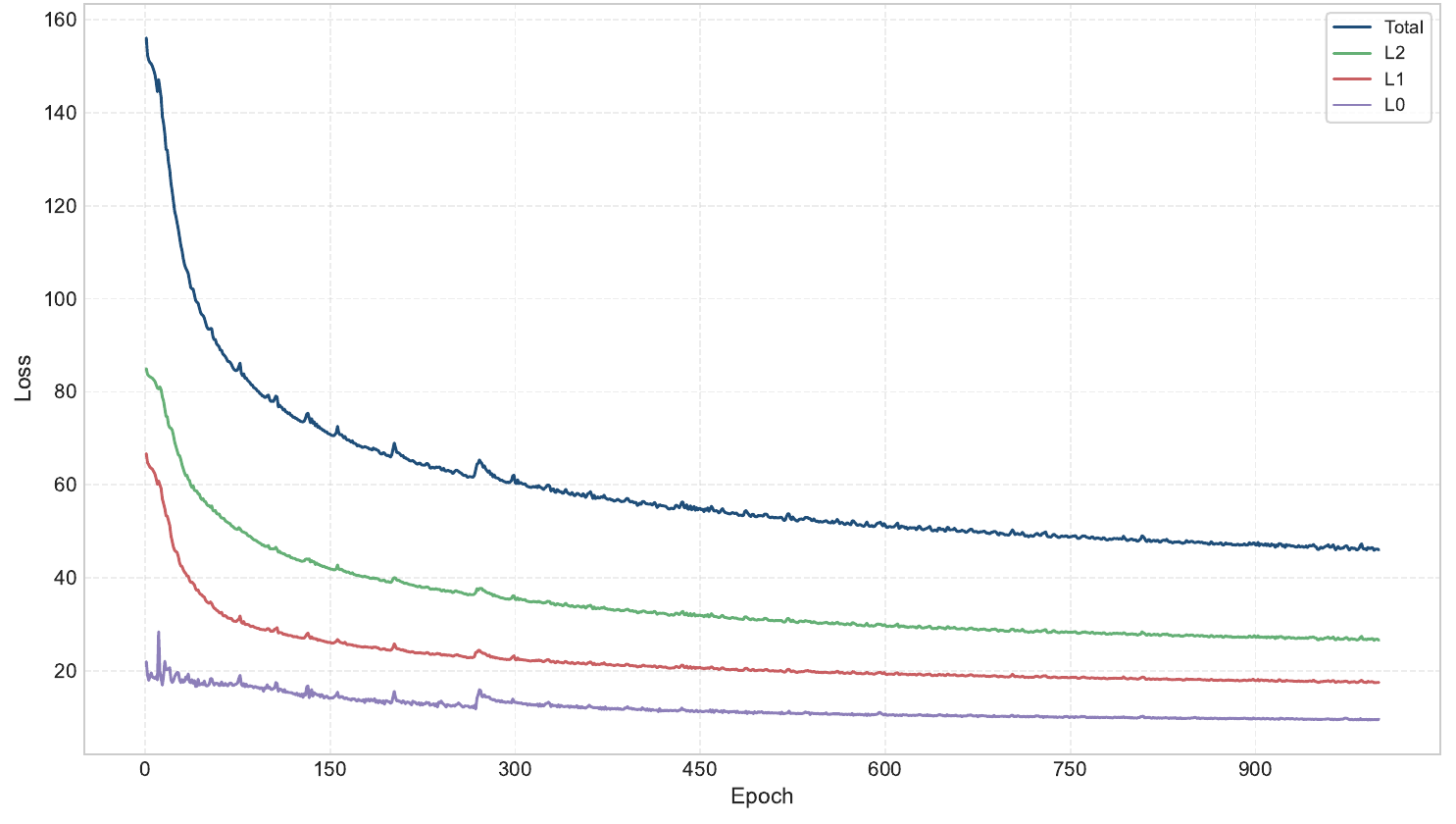}
    \caption{Loss--epoch convergence curve of the training process, including Total, L2, L1, and L0 losses. The losses stabilize after approximately 500 epochs, indicating that the model has nearly converged.}
    \label{fig:b}
\end{figure}

To determine an appropriate number of training epochs, we plot the loss--epoch convergence curve, as shown in Fig.~\ref{fig:b}. It can be observed that the training loss becomes nearly stable after approximately 500 epochs, indicating that the model has almost fully converged. Therefore, we set the number of training epochs to 500 in our experiments.

\end{document}